\documentclass{article}
\usepackage{subfigure,url}
\usepackage{spconf,amsmath,graphicx,subcaption,amsfonts,mathtools,booktabs,diagbox,tikz,xcolor,bm,colortbl,enumitem}

\usepackage{hyperref}
\usepackage{fancyhdr}
\fancypagestyle{firstpage}{
  \fancyhf{}
  \fancyfoot[C]{\footnotesize{
    © 2026 IEEE. Personal use of this material is permitted. Permission from IEEE must be obtained for all other uses, in any current or future media, including reprinting/republishing this material for advertising or promotional purposes, creating new collective works, for resale or redistribution to servers or lists, or reuse of any copyrighted component of this work in other works.
  }}
}
\newcommand{\tightbox}[2]{%
  {\setlength{\fboxsep}{0.25pt}\colorbox{#1}{#2}}%
}

\definecolor{artifact}{HTML}{FF887B}
\definecolor{zoom}{HTML}{B4E3FF}
\definecolor{Gold}{RGB}{255,215,0}
\definecolor{Silver}{RGB}{192,192,192}
\definecolor{Bronze}{RGB}{205,127,50}
\colorlet{GoldSoft}{Gold!25}
\colorlet{SilverSoft}{Silver!25}
\colorlet{BronzeSoft}{Bronze!25}
\newcommand{\best}[1]{\tightbox{GoldSoft}{#1}}
\newcommand{\second}[1]{\tightbox{SilverSoft}{#1}}
\newcommand{\third}[1]{\tightbox{BronzeSoft}{#1}}

\title{Nix and Fix: Targeting 1000× Compression of 3D Gaussian Splatting with Diffusion Models}

\name{Cem~Eteke \textsuperscript{1,2}~and~Enzo~Tartaglione \textsuperscript{2}\thanks{Corresponding author email: enzo.tartaglione@telecom-paris.fr}
}
\address{\textsuperscript{1} Chair of Media Technology, Munich Institute of Robotics and Machine Intelligence \\ School of Computation, Information, and Technology \\ Technical University of Munich, 80333 Munich, Germany \\ \textsuperscript{2}LTCI, T\'el\'ecom Paris, Institut Polytechnique de Paris, France
}

\begin{document}
% For arXiv
\thispagestyle{firstpage}

\maketitle
\begin{abstract}
3D Gaussian Splatting (3DGS) revolutionized novel view rendering. Instead of inferring from dense spatial points, as implicit representations do, 3DGS uses sparse Gaussians. This enables real-time performance but increases space requirements, hindering rate-constrained applications. 3DGS compression emerged as a field aimed at alleviating this issue. While impressive progress has been made, at low rates, compression introduces artifacts that degrade visual quality significantly. We introduce NiFi, a method for extreme 3DGS compression through restoration via artifact-aware, diffusion-based one-step distillation. We show that our method achieves state-of-the-art perceptual quality at extremely low rates, down to $0.1~\text{MB}$, and towards $1000\times$ rate improvement over 3DGS at comparable perceptual performance. Code is available at: \url{https://github.com/ceteke/nifi}
\end{abstract}

\begin{keywords}
3DGS compression, image restoration, flow matching, diffusion models
\end{keywords}
\section{Introduction}
\label{sec:intro}

Utilizing 3D Gaussian Splatting (3DGS) for novel-view rendering has emerged as an alternative to implicit neural radiance models~\cite{kerbl20233d}. Instead of the dense prediction scheme of the latter, i.e., estimating color and opacity from given points in 3D space, 3DGS fits sparse Gaussians with attributes, such as color, scale, and position, in relevant, non-empty regions. This representation, combined with parallelizable rasterization, enabled real-time novel-view rendering.

The real-time nature of 3DGS is highly relevant to applications such as immersive communications. Nevertheless, the introduction of Gaussians with attributes, i.e., primitives, increases the rate significantly compared to a low-parameter implicit neural network. To alleviate this issue, 3DGS compression has garnered significant research interest in recent years. Proposed methodologies range from unstructured approaches, such as pruning and quantization, to structured approaches, including anchor and graph-based methods~\cite{ali2025compression}.

\begin{figure}[t]
    \centering
    \includegraphics[width=0.85\columnwidth]{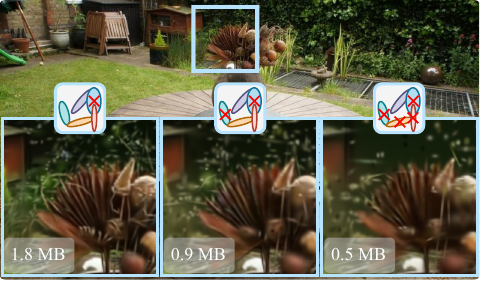}
    \caption{Artifacts resulting from 3DGS compression at different rates. Rate control is achieved through pruning ($\color{red}\bm{\times}$), quantization, and entropy coding. Notice the loss of geometry, texture, and radiance.}
    \label{fig:artifacts}
\end{figure}

Even though 3DGS compression achieves up to $100 \times$ rate gain over a baseline 3DGS~\cite{chen2025hac++}, at extremely low rates, through the degradation of the underlying 3D representation, rendering introduces complex artifacts. This is unlike 2D image degradation, such as downsampling, as it results from aggregating the distorted 3D representations. Resulting artifacts can include, but are not limited to, blur, loss, and degradation of texture, geometry, and radiance. Figure~\ref{fig:artifacts} displays examples of such rendering artifacts for a scene at extreme low rates. Restoring these artifacts would significantly improve the rendering performance, thus enabling the implementation of 3DGS in data-constrained applications. This restoration perspective for extremely low rates is underexplored.

Although deep image restoration methods have been successful in 2D restoration scenarios, as discussed, 3DGS compression artifacts pose a greater challenge. To this end, generative perceptual restoration represents a promising direction, as pretrained Latent Diffusion Models provide strong natural-image priors and achieve high perceptual quality in data-driven restoration even under severe distortion~\cite{rombach2022high}. 

We introduce NiFi, which enables extreme 3DGS compression through variational diffusion distillation for one-step restoration~\cite{yin2024one}. We perform data-driven artifact-aware distillation through a synthetic 3DGS compression dataset. Rather than directly restoring from degraded input, we reparameterize the inference by mapping the image onto an immediate diffusion step, allowing the model to exploit stochastic diversity. We compare our approach with classical~\cite{danielyan2011bm3d}, deep learning-based~\cite{liang2021swinir}, generative~\cite{lin2024diffbir}, and 3DGS restoration~\cite{wu2025difix3d+} methods. We report state-of-the-art restoration performance and an almost $1000\times$ reduction in rate, with perceptual performance comparable to the non-compressed 3DGS. Our contributions therefore are the following:
\begin{itemize}[noitemsep,topsep=0pt]
    \item We propose a compression framework for 3DGS enabling access to extremely low bitrates, where the perceptual quality can be trade-offed with decoding complexity (Sec.~\ref{sec:method})
    \item We conduct extensive comparisons against classical, deep learning-based, and generative restoration methods, demonstrating state-of-the-art  performance under extreme compression (Sec.~\ref{sec:eval}).
\end{itemize}

%To sum up, we propose an extreme 3DGS compression method via (i) 3DGS compression artifact synthesis for (ii) data-driven one-step restoring variational diffusion distillation, extended through (iii) mapping to an intermediate diffusion step for enhanced restoration quality.

\section{Related Work}

%\subsection{3D Gaussian Splatting}

\textbf{3D Gaussian Splatting.} Kerbl et al. used 3DGS to represent a scene with a set of sparse Gaussians $\mathcal{G} := \left\{G_i(x)\right\}_{i=1}^{L}$, i.e., primitives~\cite{kerbl20233d}. In this formulation, each primitive’s mean and covariance represent its three-dimensional geometry: the mean encodes position, while the covariance captures rotation and scale. Novel view rendering is achieved by ordering the primitives with respect to a given camera view and projecting them onto the image plane. The pixel color values are then rendered using $\alpha$-compositing of color and opacity, where the color is represented using view-dependent spherical harmonic coefficients. This extends the primitives to include color and opacity as additional attributes: storing these results in high space demand, hindering data-constrained applications.

\textbf{Towards More Efficient Representations.} 
% Removed references to fit to 6 pages:
 % - Papantonakis et al. introduced resolution-aware redundancy reduction to opacity-based pruning \cite{papantonakis2024reducing}.
 % - Ali et al. utilized gradient-based trimming \cite{ali2024trimming}.
 % - LP-3DGS introduced learnable pruning through a learnable binary mask, eliminating the need to design scene-specific pruning \cite{zhang2024lp}.
 % - Kim et al. extended this by using color gradients for color-informed densification \cite{kim2024color}
One dominant approach to efficient 3DGS representation is to reduce the number of primitives, namely, by pruning. Pruning methods remove primitives or regulate densification, reducing the overall size. The pruning can be explicitly carried out in accordance with various measures. For instance, CompGS utilized size and opacity~\cite{navaneet2024compgs}. In addition to these static criteria, many approaches rely on gradient-based measures to guide pruning masks. EfficientGS evaluated which primitives are close to convergence based on gradients of positions and applied an adaptive densification strategy~\cite{liu2025efficientgs}. To further support variable rates, GoDe used gradient-based pruning at multiple rates and finetuning for adaptation~\cite{di2025gode}. 

Attribute-based measures for explicit pruning, whether static (e.g., opacity) or dynamic (e.g., gradient), might not be informative enough about rendering performance. Rendering-based pruning methods address this issue. LightGaussian used a scoring-based approach for pruning primitives that do not contribute to the rendering quality~\cite{fan2024lightgaussian}. Furthermore, EAGLES introduced an influence metric to prune the primitives with low influence on the rasterization~\cite{girish2024eagles}. However, these approaches still rely on explicitly evaluating and removing individual primitives. Alternatively, structured methods such as Scaffold-GS introduce reference points, i.e., anchors, that define dynamic primitives, thereby reducing the number of primitives to store~\cite{lu2024scaffold}.

While pruning and anchoring approaches reduce the number of primitives, the redundancies can be further exploited. To that extent, scalar ~\cite{di2025gode} and vector quantization~\cite{girish2024eagles, fan2024lightgaussian, navaneet2024compgs} are combined with entropy coding. To further enable entropy models for 3DGS, HAC++ utilized mutual information among elements in a hash-grid of anchors to create a context model~\cite{chen2025hac++}. Furthermore, HEMGS introduced a learnable entropy model and a hyperprior network~\cite{liu2024hemgs}. CodecGS utilized 2D feature planes and entropy modelling to compress 3DGS with video codecs~\cite{Lee_2025_ICCV}. While the discussed approaches improve efficiency, at extremely low rates they degrade the underlying 3D representation and introduce complex artifacts. We examined such a case in Fig.~\ref{fig:artifacts}.

\begin{figure*}[t]
    \centering
    \includegraphics[width=1\textwidth]{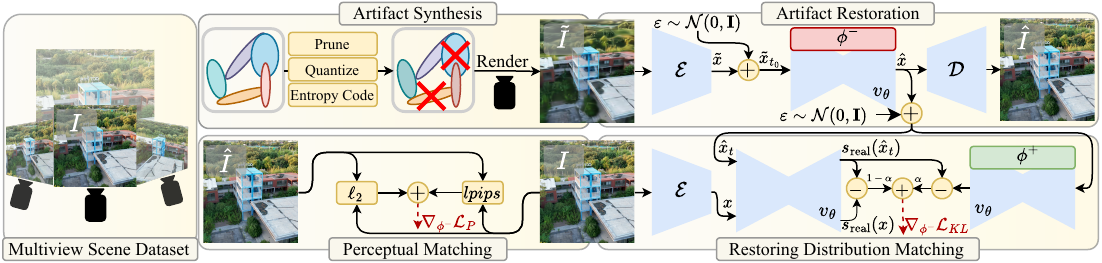}
    \caption{The overall pipeline of NiFi. We create a 3DGS restoration dataset of degraded $\tilde{I}$ and high-quality $I$ frames via \textit{Artifact Synthesis}, through pruning, quantization, and entropy coding at three rates. $\tilde{I}$ is mapped to the latent space and to an intermediate step $t_0$ in the diffusion trajectory. $\hat{I}$ is obtained in one step via adapter $\phi^-$ extending a frozen flow matching backbone $v_\theta$ for \textit{Artifact Restoration}. $\phi^-$ is trained with \textit{Restoring Distribution Matching} through the critic adapter $\phi^+$ and \textit{Perceptual Matching} between $\hat{I}$ and $I$. Only Artifact Restoration is performed during inference. The architecture of $v_\theta$ is illustrative.}
    \label{fig:pipeline}
\end{figure*}

\textbf{Image and 3D Gaussian Splatting Restoration.} Classical methods such as block matching and 3D filtering (BM3D) rely on known degradation models~\cite{danielyan2011bm3d}. Wang et al. demonstrated that data-driven training of convolutional neural networks (CNNs) using simulated degradation removes this requirement~\cite{wang2021real}. Building on this idea, SwinIR extended the approach to the Swin Transformer architecture~\cite{liang2021swinir}. More recently, diffusion models (DMs) have achieved perceptually superior restoration quality, surpassing the limits of regression-based methods~\cite{li2025diffusion}. Subsequent works have explored hybrid strategies, with DiffBIR combining non-generative restoration methods and controllable DMs~\cite{lin2024diffbir}, and TSD-SR introducing a reconstruction term into distribution-matching distillation of DMs to enable superresolution~\cite{dong2025tsd}. However effective, these approaches cannot generalize to the artifacts introduced by 3DGS, as the degradation of the underlying 3D representation yields complex artifacts beyond the scope of widely studied image restoration tasks.

DMs have been successfully extended to 3DGS novel-view artifact restoration. GSFix3D used the backward diffusion to update 3DGS primitives~\cite{wei2025gsfix3d}. Difix3D, on the other hand, trained the diffusion backbone for restoration, keeping the primitives fixed~\cite{wu2025difix3d+}. Though they enable better novel-view synthesis, these methods are sensitive to compression artifacts. To account for the artifacts, Shin et al. used a regression-based restoration combined with residual coding for 3DGS compression~\cite{shin2025leveraging}. However, residual coding increases the rate, and regression-based approaches do not achieve high perceptual quality. In our work, we introduce an extreme 3DGS compression approach that leverages the restoration capabilities of one-step diffusion distillation~\cite{dong2025tsd}.

\section{Methodology} \label{sec:method}
Information loss at the 3D representation distorts both the geometry and appearance, thus compressing 3DGS, especially at extremely low rates, produces complex artifacts. We aim to restore these artifacts by formulating a blind image restoration problem, enabling 3DGS compression at extreme-low rates. 

\textbf{Artifact Synthesis} step shown in Fig.~\ref{fig:pipeline} generates distorted and clean image pairs by simulating compression artifacts at distinct rates. This simulation approach lies at the core of learning-based blind image restoration methods~\cite{wang2021real}. To support the simulation of artifacts at a variable rate, we use GoDe~\cite{di2025gode}. We prune pretrained 3DGS model at three levels $\left\{ \mathcal{G}_0, \mathcal{G}_1, \mathcal{G}_2 \right\}$, where $|\mathcal{G}_l| = c_{min}\mathrm{exp}\left( \frac{\log(|\mathcal{G}_{L-1}|) - \log(c_{min})}{L-1} \right)$. $c_{min}$ is the minimum cardinality and $|\mathcal{G}_0|=c_{min}$. The pruning metric is the gradient of the rendering loss with respect to the attributes $\mathrm{lowest}_{|\mathcal{G}_l|} \lVert \frac{\partial \mathcal{L}}{\partial \mathcal{G}} \rVert_2$. The 3DGS model is then fine-tuned with pruned versions at three levels, and the resulting attributes are stored using 8-bit quantization and entropy coding. Given a training view $I$, we finalize the simulation by rendering degraded $\tilde{I}$ from a low-rate $\mathcal{G}_l$.

We formulate recovering the original frame $I$ given degraded $\tilde{I}$ as a distribution matching problem where we match, for a restored estimate $\hat{I}$, distributions $p_{\mathrm{restore}}$ and $p_{\mathrm{real}}$

\begin{equation} \label{eq:kl}
    \mathcal{L}_{KL} = \operatorname*{\mathbb{E}}_{\hat{I} \sim p_{\mathrm{restore}}} \left[ \log{\left( \frac{p_\mathrm{restore}(\hat{I})}{p_\mathrm{real}(\hat{I})} \right)} \right] \text{.}
\end{equation}

To this end, we leverage a rectified flow model as the image prior. We interchangeably use the general term "diffusion model" and aim to learn $p_{\mathrm{restore}}$~\cite{rombach2022high}. A latent diffusion model (LDM) consists of an encoder $\mathcal{E}$, a decoder $\mathcal{D}$, and a velocity model $v_\theta$. The LDM serves as the generative latent image distribution by parameterizing the forward process

\begin{equation} \label{eq:forward}
    x_t = \left(1-t\right)x + t\varepsilon,
\end{equation} where $\varepsilon \sim \mathcal{N}(0, \mathbf{I})$. The model $v_\theta$ parameterizes the velocity field $\frac{dx}{dt}=v_\theta(x_t,t)$. The inverse process is performed by integrating the discretized ODE via indices $i\in \{T,\dots,1\}$

\begin{equation}
    x_{\frac{i-1}{T}} = x_{\frac{i}{T}} - \left(\frac{i}{T} - \frac{i-1}{T}\right)v_{\theta}\left(x_{\frac{i}{T}}, \frac{i}{T}\right) \text{.}
\end{equation}

\textbf{Artifact Restoration} extends $v_\theta$ with a low-rank adapter $\phi^-$ to obtain the estimated clean image $\hat{I}$ via the resulting one-step restoration from the model $v_{\theta, \phi^-}$. Furthermore, rather than directly restoring from $\tilde{x}$, we project the latents to an intermediate time $t_0$ using the forward model in Eq.~\ref{eq:forward}, allowing the model to exploit diversity in a one-step formulation: $\hat{x} = \tilde{x}_{t_0} - t_0 v_{\theta,\phi^-}(\tilde{x}_{t_0}, t_0)$.

\textbf{Restoration Distribution Matching} step, illustrated in Fig.~\ref{fig:pipeline}, minimizes the KL-Divergence term in Eq.~\ref{eq:kl} with respect to $\phi^-$. To that extent, the variational distillation formulation approximates the intractable $\nabla_{\phi^-}\mathcal{L}_{KL}$ through

\begin{table*}[t]
\caption{Quantitative results at three rates and the single-rate 3DGS baseline. For LPIPS and DISTS, lower is better ($\downarrow$). We highlight \best{\textbf{best}}, \second{second-best} and \third{third-best} performing methods.}
\centering
\scriptsize{
\begin{tabular}{l@{\hspace{1pt}}l@{\hspace{2pt}}ccccccccc}
\cline{2-11}
 &
  \multicolumn{1}{c}{} &
  \multicolumn{3}{c}{\begin{tabular}[c]{@{}c@{}}Mip-NeRF360 \cite{barron2022mip}\\ (LPIPS $\downarrow$ / DISTS $\downarrow$)\end{tabular}} &
  \multicolumn{3}{c}{\begin{tabular}[c]{@{}c@{}}Tanks \& Temples \cite{knapitsch2017tanks}\\ (LPIPS $\downarrow$ / DISTS $\downarrow$)\end{tabular}} &
  \multicolumn{3}{c}{\begin{tabular}[c]{@{}c@{}}DeepBlending \cite{hedman2018deep}\\ (LPIPS $\downarrow$ / DISTS $\downarrow$)\end{tabular}} \\ \cline{2-11} 
 & Size (MB)   & \multicolumn{3}{c}{576}                       & \multicolumn{3}{c}{339}                       & \multicolumn{3}{c}{555}                       \\
 & 3DGS-30K \cite{kerbl20233d}    & \multicolumn{3}{c}{0.156 / 0.078}             & \multicolumn{3}{c}{0.125 / 0.067}             & \multicolumn{3}{c}{0.125 / 0.098}             \\ \cline{2-11} 
 & Size (MB)   & 1.152         & 0.357         & 0.223         & 1.312         & 0.381         & 0.230         & 0.599         & 0.183         & 0.110          \\
 & HAC++ \cite{chen2025hac++}      & 0.285 / 0.158 & 0.409 / 0.201 & 0.455 / 0.220 & 0.210 / 0.113 & 0.317 / 0.169 & 0.364 / 0.193 & 0.201 / 0.161 & 0.282 / 0.215 & 0.337 / 0.249 \\ \cline{3-11} 
 & $+$ BM3D \cite{danielyan2011bm3d}    & 0.308 / 0.164 & 0.423 / 0.206 & 0.466 / 0.226 & 0.225 / 0.124 & 0.325 / 0.175 & 0.371 / 0.199 & 0.209 / 0.172 & 0.286 / 0.222 & 0.340 / 0.254 \\
 & $+$ SwinIR \cite{liang2021swinir}  & 0.346 / 0.218 & 0.436 / 0.239 & 0.468 / 0.248 & 0.281 / 0.159 & 0.350 / 0.192 & 0.381 / 0.209 & 0.236 / 0.180 & 0.290 / 0.213 & 0.333 / 0.237 \\
 & $+$ Img2Img \cite{esser2024scaling} & 0.341 / 0.156 & 0.438 / 0.200 & 0.477 / 0.220 & 0.299 / 0.128 & 0.372 / 0.174 & 0.408 / 0.197 & 0.246 / 0.169 & 0.313 / 0.218 & 0.362 / 0.249 \\
 & $+$ DiffBIR \cite{lin2024diffbir} & 0.350 / 0.178 & 0.417 / 0.203 & 0.446 / 0.215 & 0.283 / 0.154 & 0.338 / 0.179 & 0.375 / 0.196 & 0.298 / 0.208 & 0.368 / 0.228 & 0.410 / 0.241 \\
 & $+$ Difix3D \cite{wu2025difix3d+} & \third{0.238} / \third{0.133} & \third{0.300} / \second{0.145} & \third{0.330} / \second{0.152} & \third{0.165} / \second{0.088} & \third{0.235} / \second{0.116} & \third{0.272} / \second{0.131} & \second{0.158} / \second{0.111} & \second{0.216} / \second{0.143} & \second{0.257} / \second{0.161} \\
 &
  $+$ NiFi (Ours) &
  \best{\textbf{0.178}} / \best{\textbf{0.109}} &
  \best{\textbf{0.235}} / \best{\textbf{0.133}} &
  \best{\textbf{0.265}} / \best{\textbf{0.153}} &
  \best{\textbf{0.128}} / \best{\textbf{0.076}} &
  \best{\textbf{0.180}} / \best{\textbf{0.095}} &
  \best{\textbf{0.212}} / \best{\textbf{0.109}} &
  \best{\textbf{0.133}} / \best{\textbf{0.101}} &
  \best{\textbf{0.180}} / \best{\textbf{0.131}} &
  \best{\textbf{0.218}} / \best{\textbf{0.156}} \\ \cline{3-11} 
 &
  \begin{tabular}[c]{@{}l@{}}$+$ NiFi (w/o $t_0$)\end{tabular} &
  \second{0.211} / \second{0.129} &
  \second{0.287} / \third{0.174} &
  \second{0.324} / \third{0.197} &
  \second{0.153} / \third{0.104} &
  \second{0.229} / \third{0.147} &
  \second{0.269} / \third{0.173} &
  \third{0.162} / \third{0.147} &
  \third{0.231} / \third{0.193} &
  \third{0.282} / \third{0.223} \\ \cline{2-11} 
\end{tabular}
\label{tab:quant}
}
\end{table*}

\begin{equation}
    \nabla_{\phi^-}\mathcal{L}_{KL} = \operatorname*{\mathbb{E}}_{\hat{x},t} \left[ \left( s_{\mathrm{restore}}(\hat{x}) - s_{\mathrm{real}}(\hat{x}) \right)\frac{\partial\hat{x}}{\partial \phi^-}   \right]
\end{equation} where $s(x) \coloneq \nabla_x \log p(x)$ is the score, approximated from the noise-free estimate of a diffusion model~\cite{yin2024one}. For the real image distribution, this is $s_{\mathrm{real}}(\hat{x}) \approx \hat{x}_t -t v_\theta(\hat{x}_t, t) $, where $t \sim \mathcal{U}[\frac{t_{min}}{T}, \frac{t_{max}}{T}]$, sampled during training. The restoration distribution $p_{\mathrm{restore}}$ is modeled with another low-rank adapter that extends frozen $v_\theta$ with parameters $\phi^{+}$. Hence, $s_{\mathrm{restore}}(\hat{x}) \approx \hat{x}_t - t v_{\theta, \phi^+}(\hat{x}_t, t)$. The model $v_{\theta, \phi^+}$ enables matching by learning the distribution of the restored frames.

Though effective for single-step synthesis, this formulation alone does not explicitly guide distillation to restore a high-quality frame while preserving the fidelity. To achieve that, we utilize guidance in the ground-truth direction~\cite{dong2025tsd}. Dong et al. extended the distribution matching as

\begin{equation}
\begin{aligned}
\nabla_{\phi^-}\mathcal{L}_{\mathrm{KL}}
&= \alpha \operatorname*{\mathbb{E}}_{\hat{x},t} \left[ \left( s_{\mathrm{restore}}(\hat{x}) - s_{\mathrm{real}}(\hat{x})  \right) \frac{\partial\hat{x}}{\partial \phi^-} \right] + \\
& + (1-\alpha)\,
\operatorname*{\mathbb{E}}_{x,\hat{x},t}
\Big[
\left(s_{\mathrm{real}}(x)
- s_{\mathrm{real}}(\hat{x})
\,\right)\frac{\partial \hat{x}}{\partial \phi^-}
\Big] \text{.}
\end{aligned}
\end{equation}

We leverage the high-quality ground $x$ available through the simulated artifacts. This enables a score-matching term between $x$ and $\hat{x}$. This formulation has been successful in image superresolution \cite{dong2025tsd}. However, as discussed, the 3DGS compression artifacts are complex. To circumvent this issue, we introduced restoration at $t_0$. 

\textbf{Perceptual Matching} step, on top of the restoring distribution matching formulation, utilizes $\ell_2$ and $lpips$ loss functions to minimize the perceptual loss. The final optimization objective of $\phi^-$ becomes

\begin{equation} \label{eq:psi-}
    \mathcal{L}_{\phi^-} = \mathcal{L}_{KL}~+~\ell_2(x,~\hat{x})~+~lpips(x,~\hat{x})\text{.}
\end{equation}

During inference, only the artifact restoration step is performed. For an image $\tilde{I}$ with latents $\tilde{x} = \mathcal{E}(\tilde{I})$ the restored latents $\hat{x}$ are estimated as

\begin{equation} \label{eq:inference}
    \hat{x} = \tilde{x}_{t_0} - t_0 v_{\theta,\phi^-}(\tilde{x}_{t_0},~t_0)
\end{equation} where $\tilde{x}_{t_0}$ is the forward model in Eq.~\ref{eq:forward} and $t_0$ is a hyperparameter that specifies the position on the diffusion trajectory for restoration of the complex 3DGS artifacts, enabling high perceptual performance.

The parameters $\phi^+$ are trained to minimize the following flow loss term where $t \sim \mathcal{U}[\frac{t_{min}}{T}, \frac{t_{max}}{T}]$, $\varepsilon \sim \mathcal{N}(0,\mathbf{I})$, and $\hat{x}_t$ is the forward model in Eq.~\ref{eq:forward} with $\hat{x}_1 = \varepsilon$ applied to the predicted restoration

\begin{equation} \label{eq:psi+}
    \mathcal{L}_{\phi^+} = \ell_2(\varepsilon - \hat{x},~v_{\theta,\phi^+}(\hat{x}_t,~t)) \text{.}
\end{equation}

The training oscillates between optimizing for $\mathcal{L}_{\phi^-}$ and $\mathcal{L}_{\phi^+}$, updating the parameters $\phi^-$ and $\phi^+$ respectively. The first restores 3DGS compression artifacts by matching the distribution of natural images and perceptual fidelity, while the latter enables modeling the restored image distribution for distribution-matching distillation. We name our overall approach NiFi.

\section{Experiments}

\subsection{Implementation Details}

We used the DL3DV dataset with $10^3$ scenes to create the simulated 3DGS compression artifacts dataset~\cite{ling2024dl3dv}. We set the minimum number of primitives for pruning $c_{min}=4096$ and selected the number of primitives at three rates as described in Sec.~\ref{sec:method}. We trained the two low-rank adapters, $\phi^-$ and $\phi^+$ with rank 64, on the transformer-based backbone of Stable Diffusion 3 (SD3)~\cite{esser2024scaling}. We used classifier-free guidance to compute $s_{\mathrm{real}}$ and $s_{\mathrm{restore}}$ with a guidance scale of $7.5$. The parameters $\phi^-$ and $\phi^+$ are updated one after another to minimize the objectives presented in Eq.~\ref{eq:psi-} and Eq.~\ref{eq:psi+}, respectively. To optimize these parameters, we used AdamW with learning rates of $5\times10^{-6}$ and $10^{-6}$, weight decay of $10^{-4}$, and gradient clipping of $1.0$. We followed Dong et al. in setting $\alpha=0.7$ and Difix3D in setting $t_0=\frac{199}{1000}$. We selected a random training view from each scene as each element of a minibatch, rather than randomly selecting from the set of all frames; hence, an epoch consisted of $10^3$ steps. We trained for $60\times10^3$ steps with a batch size of $4$ on an NVIDIA H200 GPU. This training took approximately 2 days. As the backbone is a text-to-image model, we used Qwen2.5-VL to extract prompts from the first training frame of each scene~\cite{Qwen2.5-VL}. We disabled these prompts with a probability of $\frac{1}{10}$ during training. In practice, we included low-rank adapter weights of the encoder $\mathcal{E}$ in the set $\phi^-$.

\begin{figure*}[t]
    \centering
    \subfigure{
        \includegraphics[width=0.492\textwidth]{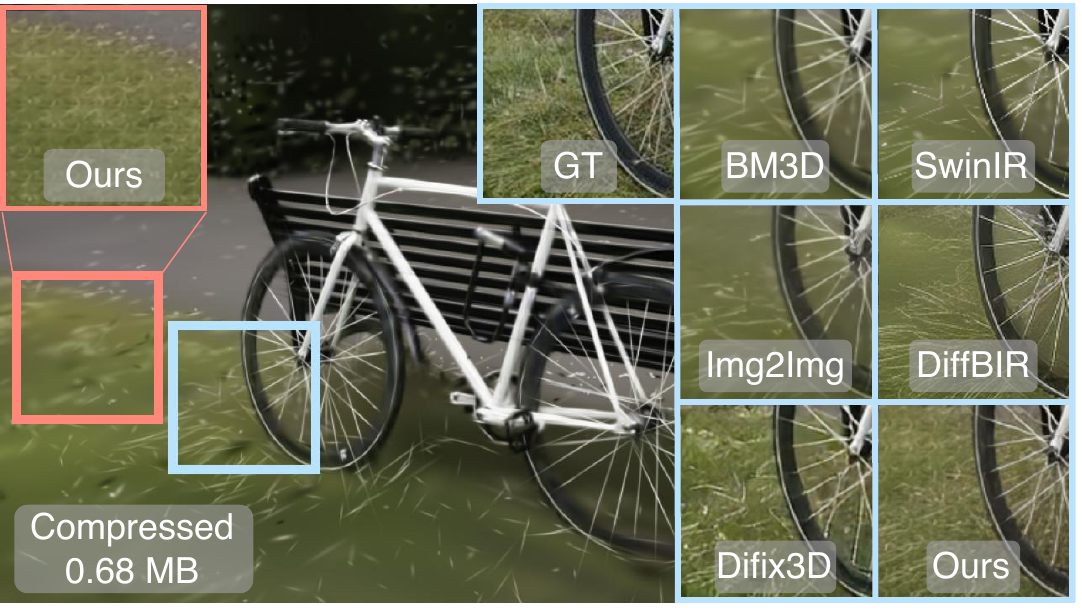}
    }\hspace{-0.7em}
    \subfigure{
        \includegraphics[width=0.492\textwidth]{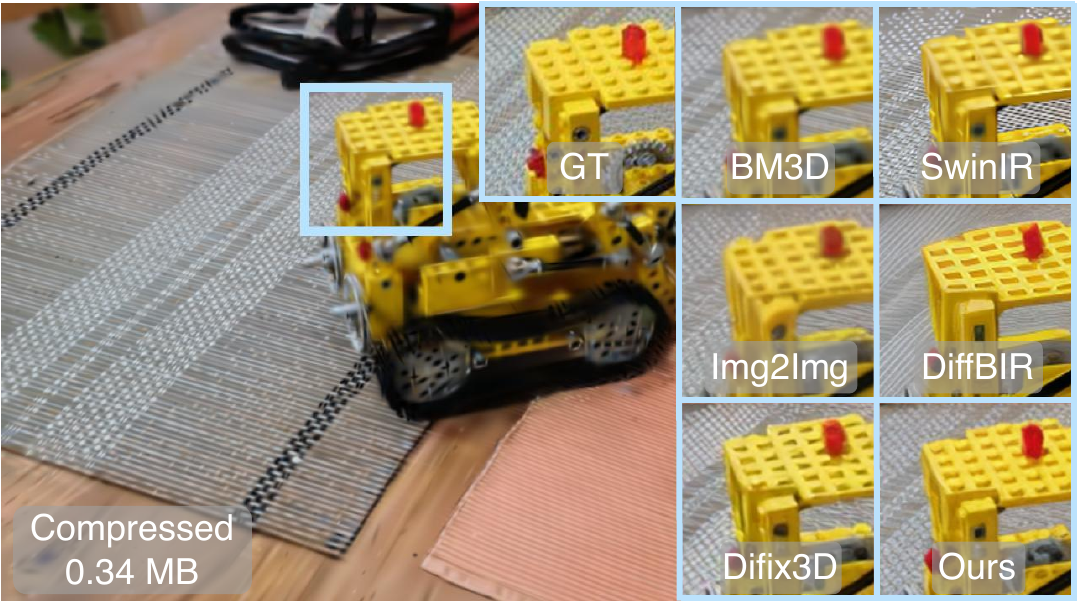}
    }\\
    \vspace{-1.1em}
    \subfigure{
        \includegraphics[width=0.492\textwidth]{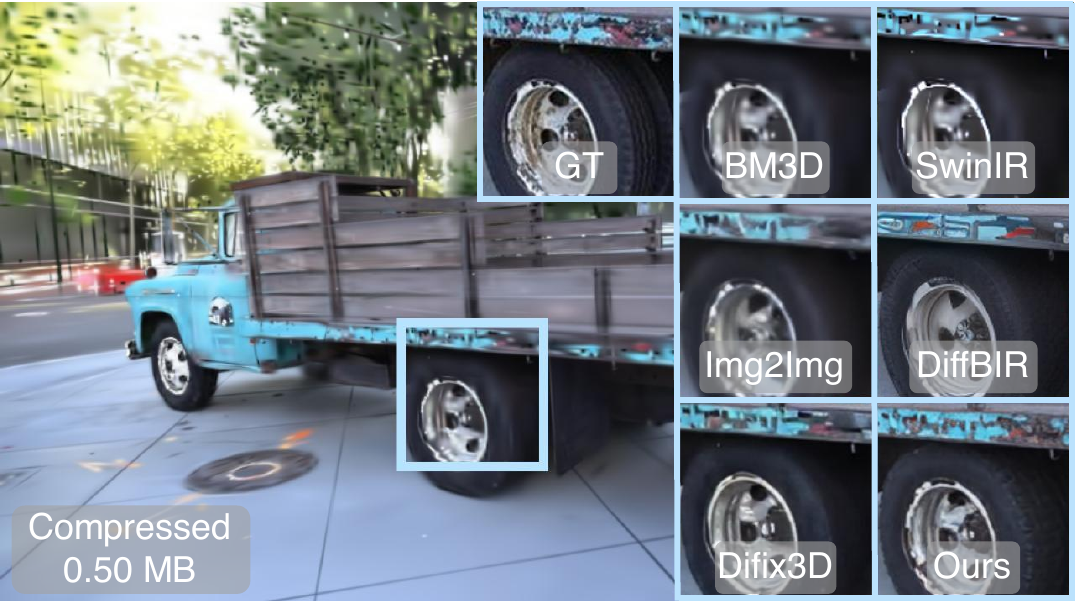}
    }\hspace{-0.7em}
     \subfigure{
        \includegraphics[width=0.492\textwidth]{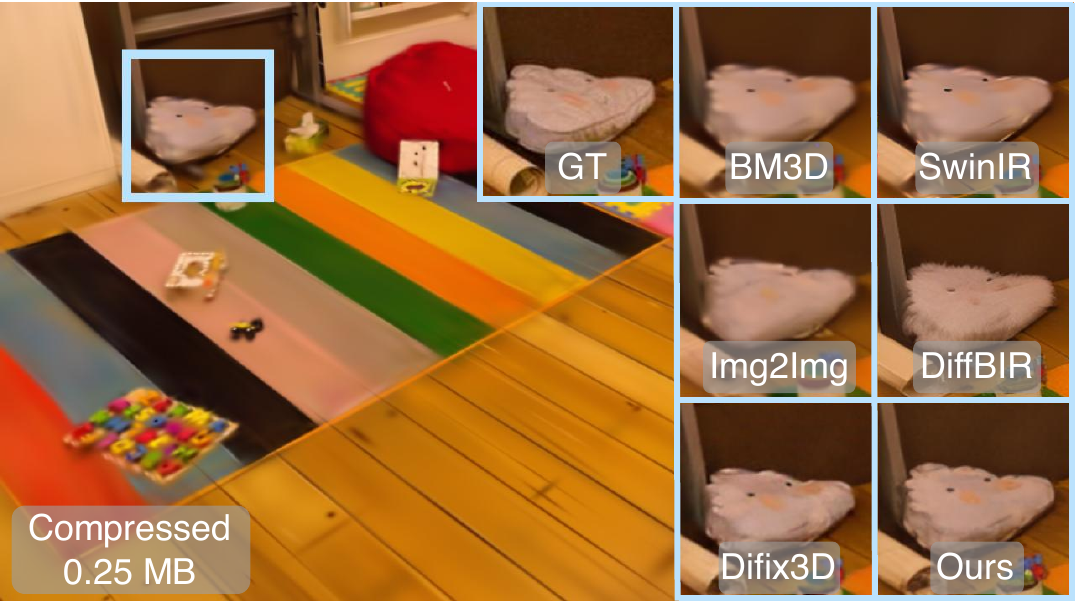}
    }
    \caption{Qualitative results of \textit{bicycle}, \textit{kitchen}, \textit{truck}, and \textit{playroom} scenes. Backgrounds are compressed 3DGS renders with overlayed restoration results compared within the \fcolorbox{zoom}{white}{\strut highlighted areas}. The additional \fcolorbox{artifact}{white}{\strut highlighted area} in the bicycle scene shows the overemphasized high-frequency components that are introduced by our method.}
    \label{fig:qual}
\end{figure*}

\subsection{Evaluation} \label{sec:eval}

We used three evaluation datasets: Mip-NeRF360~\cite{barron2022mip}, Tanks \& Temples~\cite{knapitsch2017tanks}, and DeepBlending~\cite{hedman2018deep}. We utilized HAC++ at three extremely low rates obtained through setting the rate parameter $\lambda \in \{0.1, 0.5, 1.0\}$~\cite{chen2025hac++}. We opted to use HAC++ because it is the state-of-the-art 3DGS compression method and was not used in training. Our method does not depend on this choice. We executed the restoration as in Eq.~\ref{eq:inference}, given compressed renders from novel views. We compared our approach, NiFi, with baselines of classical restoration: BM3D~\cite{danielyan2011bm3d}, deep learning-based restoration: SwinIR~\cite{liang2021swinir}, diffusion-based restoration: DiffBIR~\cite{lin2024diffbir}, and 3DGS restoration: Difix3D~\cite{wu2025difix3d+}. As an ablation study, we report results without $t_0$ and utilize the common one-step-diffusion-model practice by setting $t_0=1$, i.e., $\hat{x}=\tilde{x} - v_{\theta,\phi^-}(\tilde{x},~T)$. We furthermore report a training-free approach, Img2Img, that utilizes SD3 to denoise from $\tilde{x}_{t_0}$. As we do not aim for high perceptual quality, we report the metrics: LPIPS and DISTS.

\subsection{Results \& Discussion}

We present the quantitative results, along with the baseline uncompressed 3DGS-30K in Tab.~\ref{tab:quant}. For DeepBlending, at comparable perceptual quality to 3DGS-30K, our method achieves a rate reduction from $555$~MB to $0.599$~MB, corresponding to a $927\times$ compression. Furthermore, the results show that our method significantly improves perceptual performance at lower rates, extending the unrestored HAC++ baseline's operation to extremely low rates. Furthermore, we outperform the image restoration baselines, confirming that our approach is more effective against 3DGS compression artifacts and a 3DGS restoration baseline. Finally, intermediate mapping to $t_0$ significantly improves performance, as shown in the ablation study. These results show that, combined with HAC++, our approach enables high perceptual performance at extreme rates. The restoration takes $100$ ms without any optimizations on an NVIDIA A40 GPU.

We further present qualitative results in Fig.~\ref{fig:qual}. These qualitative results show that our method preserves scene details with perceptual quality superior to that of the baselines. However, we would also like to point out that the restoration artifacts that arise from overemphasizing high-frequency regions. We observe this phenomenon especially in fine-grained regions, such as the highlighted grassy area in Fig.~\ref{fig:qual}.

\section{Conclusion}

We introduced NiFi, an extreme 3DGS compression method that extends variational diffusion distillation for restoring 3DGS compression artifacts, enabling 3DGS compression at extremely low rates, reaching $0.110$ MB. We also demonstrated that mapping to an immediate point on the diffusion trajectory significantly improves perceptual performance. Our method achieved state-of-the-art perceptual performance compared to the baselines at such low rates and approached a $1000\times$ rate improvement over an uncompressed 3DGS. One limitation of our work is the overemphasis on high-frequency regions, especially in areas with many details, such as the grass in the bicycle scene in Fig.~\ref{fig:qual}. Our future work will focus on explaining and alleviating this issue and minimizing decoding complexity.

\bibliographystyle{IEEEbib}
\small{
\bibliography{refs}
}

\end{document}